\documentclass{article}

\PassOptionsToPackage{numbers, compress}{natbib}
\usepackage[final,sglblindworkshop]{neurips_2026}
\usepackage[utf8]{inputenc}
\usepackage[T1]{fontenc}
\usepackage{hyperref}
\usepackage{url}
\usepackage{booktabs}
\usepackage{tabularx}
\usepackage{graphicx}
\usepackage{amsfonts}
\usepackage{amsmath}
\usepackage{nicefrac}
\usepackage[expansion=false]{microtype}
\usepackage{xcolor}
\usepackage{placeins}
\usepackage{float}

\workshoptitle{AXIOM 2026: Foundations of Efficient Deep Learning}

\title{Detecting Unreliable Classifications from Quantized Models}

\author{%
  Ananya Kotla \\
  \And  
  Jorge T. Gomez \\
}

\begin{document}

\setlength{\textfloatsep}{8pt plus 2pt minus 2pt}
\setlength{\floatsep}{6pt plus 2pt minus 2pt}
\setlength{\intextsep}{8pt plus 2pt minus 2pt}
\setlength{\abovecaptionskip}{4pt}
\setlength{\belowcaptionskip}{2pt}

\maketitle

\begin{abstract}
Post-training INT8 quantization is typically validated by checking accuracy retention, which requires ground-truth labels often unavailable in low-resource clinical settings. We show a label-free alternative: comparing a model's own output before and after quantization. We compute cosine similarity between the FP32 and INT8 models' output probability vectors for each prediction, which requires only a single forward pass. This output-agreement similarity separates lower-accuracy predictions across four evaluated architectures (ResNet-50, VGG-16-BN, Wide ResNet-50, and ViT-B/16) under fixed-percentile splits, outperforming a confidence-only baseline for all four. These results show that quantized models can be compared to their baselines to assess their drift and determine whether they are well suited for clinical deployment. 

\end{abstract}

\section{Introduction}
\label{sec:intro}


Model compression techniques such as quantization reduce memory requirements and energy consumption, allowing models to run on less capable hardware. This is especially relevant for edge deployment in low-resource settings, which are becoming further constrained by the rising cost of hardware \citep{WSTS2026} and the growing requirements of deep learning (DL) models. Clinical settings are commonly resource constrained, outside of highly specialized and high-tier hospitals, and commonly require using post-training quantization \citep{DBLP:journals/corr/LitjensKBSCGLGS17, RiveraGuzman2026}. 

However, while quantized models often perform at or below the baseline set by their non-quantized counterparts, especially strong quantization such as INT8 and INT4 \citep{Nahshan2019}, the size of this gap is often unknown and varies per model and setting. Understanding this gap is important for clinical deployment, but doing so usually requires labeled test data, which can be difficult to obtain for medical applications. Expert annotation are often required, which is expensive and time consuming, particularly for segmentation tasks, and such annotations are frequently unavailable in low-resource settings.

In this work, we show that for a mammography dataset \citep{Lee2017} we can leverage the models own output before and after quantization to identify when the quantized model drifts too far from the non-quantized model leading to less reliable predictions. This can alleviate the need for annotated data and help clinicians understand if the quantized models meets their deployment needs without having to annotate any additional data. 

\textbf{Our contributions} are: 1) Exploring the impact of quantization of three different convolution based models (ResNet-50, VGG-16-BN, and Wide ResNet-50) and a transformer based model (ViT-B/16) on the publicly available mammography dataset CBIS-DDSM for cancer detection 2) Using cosine similarity of the models output for failure case detection using the available test data, 3) Demonstration that this method works across different architectures, and seeds.

\section{Related Work}
\label{sec:related}

Quant-XAI \citep{QuantXAI2026} introduced ``explainability drift,'' relying on ground-truth evaluation and a fake-quantization proxy; we extend this with a backprop-free, label-free per-case signal on the true converted model. Catalan-Tatjer et al.\ \citep{CatalanTatjer2025} showed LLM quantization robustness is governed by training dynamics. Saliency-guided QAT \citep{rezabeyk2024saliencyassistedquantizationneural} intervenes during training instead. Cross-model disagreement is a recognized label-free signal elsewhere \citep{pmlr-v202-deng23f, gorbett2026crossmodeldisagreementlabelfreecorrectness, shu2025uncoveringoverconfidentfailurescxr}; we extend this to quantization, across architectures.

\section{Method}
\label{sec:method}

\textbf{Dataset.} We use CBIS-DDSM \citep{Lee2017}, a curated mammography dataset with radiologist-annotated tumor masks, for binary classification of each breast mass or calcification as \emph{malignant} or \emph{benign}. We construct a patient-grouped split using the official train/test designation, removing 31 patients appearing in both splits, yielding 2,780 training images (2,417 train / 363 validation) and 704 held-out test images. Of these, 384 have an available ROI mask and are used for experimental analyses; classification metrics use the full 704.
 
\textbf{Model, training, and quantization.} We train and evaluate ResNet-50 \citep{He2016}, VGG-16-BN \citep{Simonyan2014VeryDC}, Wide ResNet-50 \citep{DBLP:journals/corr/ZagoruykoK16}, and ViT-B/16 \citep{Dosovitskiy2021}. All models are pre-trained on ImageNet-1K, with a randomly initialized four-layer classification head added for the final malignant-vs-benign prediction. To reduce overfitting, we freeze the earliest layers of each backbone: \texttt{conv1}, \texttt{bn1}, \texttt{layer1}, and \texttt{layer2} for ResNet-50 and Wide ResNet-50; the first 13 convolutional layers for VGG-16-BN; and the patch embedding together with the earliest transformer encoder blocks for ViT-B/16. We apply post-training static INT8 quantization via PyTorch's FX graph API (\texttt{qnnpack} backend, 500 calibration images), using the fully converted INT8 model for all reported results. Full training hyperparameters are reported in Appendix~\ref{sec:hyperparams}. All four architectures undergo the same per-case statistical analysis in Sections~\ref{sec:outputvector}--\ref{sec:baseline}; ViT-B/16's per-bin trend is reported separately in Appendix~\ref{sec:extended-architectures}.

\textbf{Model output disagreement.} For each test image, we compute cosine similarity between the FP32 and INT8 models' softmax output probability vectors, requiring one forward pass per precision per image:
\begin{equation}
\label{eq:cosine-sim}
\text{sim}_{\cos}(p_{\text{FP32}}, p_{\text{INT8}}) = \frac{\sum_i p_{\text{FP32},i} \, p_{\text{INT8},i}}{\sqrt{\sum_i p_{\text{FP32},i}^2} \sqrt{\sum_i p_{\text{INT8},i}^2}}
\end{equation}
where $p_{\text{FP32}}, p_{\text{INT8}} \in \mathbb{R}^2$ are the output probability vectors for the same input under each precision. We classify a case as \emph{reorganized} if $\text{sim}_{\cos}$ falls below $0.8$; for the full accuracy-vs-similarity trend (Figure~\ref{fig:binned-accuracy}, Appendix~\ref{sec:extended-architectures}), we additionally bin predictions into four similarity ranges ($<0.80$, $0.80$--$0.90$, $0.90$--$0.95$, $0.95$--$1.00$). All analyses use 3 seeds and the full 384-image mask-available test set (1,152 predictions per regimen).
 

\section{Experiments}
\label{sec:experiments}
\FloatBarrier

\subsection{FP32 and INT8 Performance}
\label{sec:main-result}

Evaluating FP32 vs.\ fully converted INT8 ResNet-50 across 3 seeds ($n=1{,}152$): accuracy drops from $73.0\%$ to $66.5\%$ ($-6.5$ pts). Aggregate checks alone may miss sample-level performance drops.

Table~\ref{tab:main-results} reports the aggregate FP32-vs-INT8 comparison (Accuracy, AUC, F1) for all four architectures. FP32 baselines are close across all architectures (ResNet-50: 73.0, VGG-16-BN: 72.4, ViT-B/16: 71.7, Wide ResNet-50: 69.5), with standard deviation overlapping across all models. The impact of INT8 quantization varies per model: ResNet-50 and ViT-B/16 drop about 6\% in accuracy, while VGG-16-BN and Wide ResNet-50 show only a 1\%--2\% decrease. AUC drops are architecture dependent rather than uniform: 2\% for VGG-16-BN, 5\% for ResNet-50 and ViT-B/16, and 8\% for Wide ResNet-50. F1 score drops substantially for ViT-B/16 and Wide ResNet-50 specifically, suggesting these two models would benefit from recalibration after quantization.

\begin{table}[ht]
  \caption{FP32-vs-INT8 comparison, mean $\pm$ std across 3 seeds ($n=384$ per seed), full 704-image test set.}
  \label{tab:main-results}
  \centering
  \resizebox{\textwidth}{!}{%
  \begin{tabular}{lcccccc}
    & \multicolumn{3}{c}{\textbf{FP32}} & \multicolumn{3}{c}{\textbf{INT8}} \\
    \cmidrule(lr){2-4} \cmidrule(lr){5-7}
    \textbf{Architecture} & \textbf{Acc} & \textbf{AUC} & \textbf{F1} & \textbf{Acc} & \textbf{AUC} & \textbf{F1} \\
    \midrule
    ResNet-50         & $73.0{\pm}3.1$ & $0.800{\pm}0.002$ & $66.9{\pm}2.5$ & $66.5{\pm}1.7$ & $0.751{\pm}0.012$ & $63.0{\pm}1.3$ \\
    VGG-16-BN         & $72.4{\pm}1.0$ & $0.784{\pm}0.005$ & $65.7{\pm}1.7$ & $70.3{\pm}2.6$ & $0.768{\pm}0.006$ & $62.2{\pm}3.3$ \\
    Wide ResNet-50  & $69.5{\pm}2.4$ & $0.784{\pm}0.002$ & $65.1{\pm}1.5$ & $68.5{\pm}0.7$ & $0.721{\pm}0.020$ & $51.5{\pm}8.5$ \\
    ViT-B/16          & $71.7{\pm}2.1$ & $0.772{\pm}0.005$ & $61.3{\pm}4.6$ & $65.3{\pm}4.0$ & $0.732{\pm}0.034$ & $23.7{\pm}22.1$ \\
    \bottomrule
  \end{tabular}%
  }
\end{table}

\begin{figure}[!ht]
  \centering
  \includegraphics[width=1.0\textwidth]{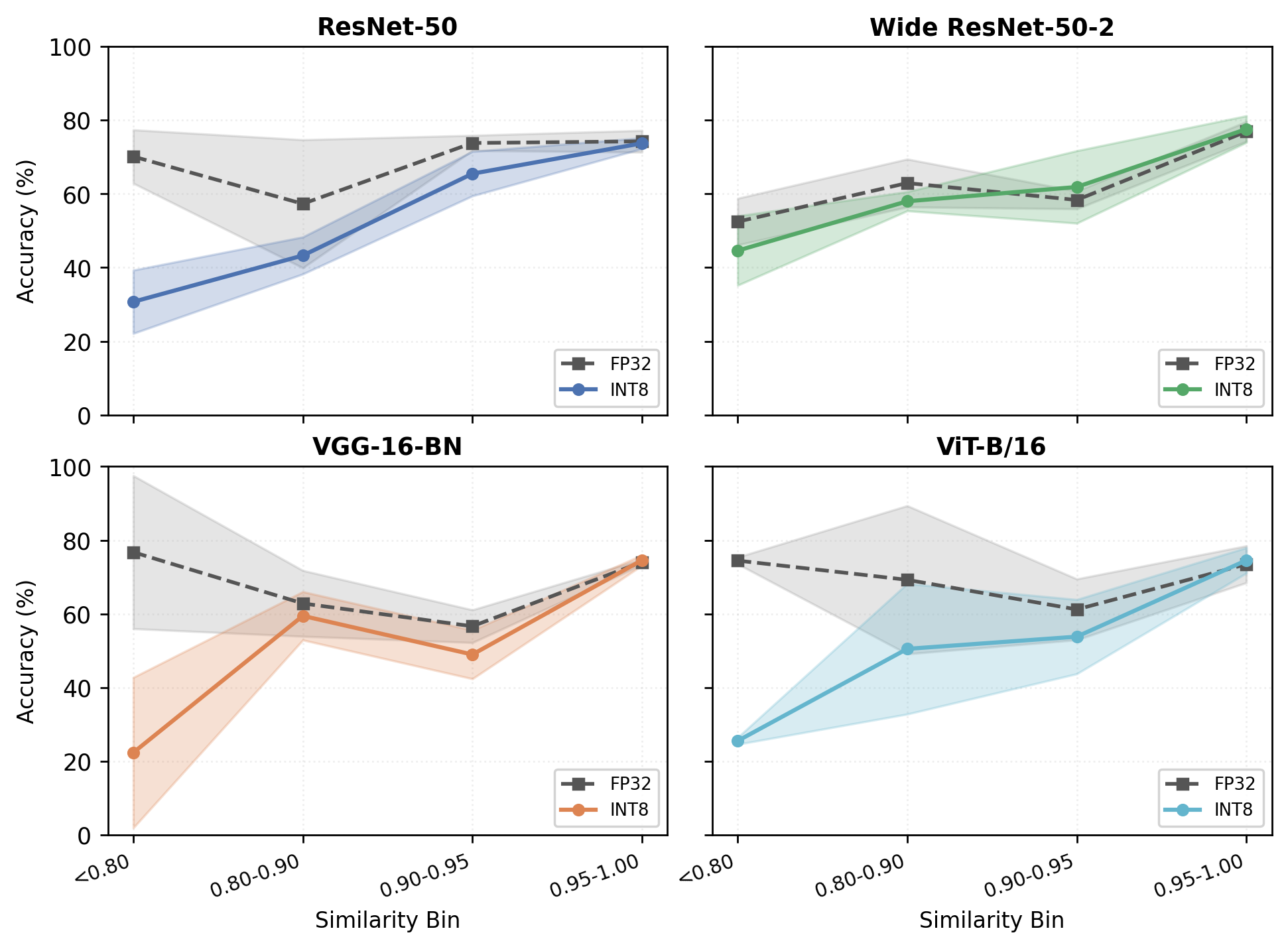}
  \caption{FP32 and INT8 accuracy and standard deviation binned by the FP32-vs-INT8 output embeddings cosine similarity for ResNet-50, Wide ResNet-50, VGG-16-BN, and ViT-B/16.}
  \label{fig:binned-accuracy}
\end{figure}

\subsection{Output Disagreement}
\label{sec:outputvector}

Stratification by output similarity is significant for all four architectures at threshold 0.8 (predictions pooled across all 3 seeds, $n=1{,}152$ per architecture): ResNet-50 ($n=123$, $30.1\%$ vs.\ $70.8\%$), VGG-16-BN ($n=56$, $28.6\%$ vs.\ $72.4\%$), Wide ResNet-50 ($n=211$, $44.5\%$ vs.\ $73.9\%$), ViT-B/16 ($n=114$, $26.3\%$ vs.\ $69.6\%$).
 
Figure~\ref{fig:binned-accuracy} shows this per seed before pooling: accuracy binned by similarity within each seed, then averaged. ResNet-50's 4 bins all contain data from all 3 seeds and follow a monotonic trend ($30.7\%\pm8.6\%$ lowest bin vs.\ $73.7\%\pm1.5\%$ highest); Wide ResNet-50, VGG-16-BN, and ViT-B/16 follow a similar trend with more per-seed noise in intermediate bins. 
 
Per-seed threshold sensitivity confirms this holds for all four architectures in every seed at threshold 0.8 (Appendix~\ref{sec:perseed-appendix}), though VGG-16-BN's smallest seed (1207, $n=4$) and ViT-B/16's smallest seed (1207, $n=4$) both reflect a genuinely lower baseline disagreement rate under static quantization relative to the other architectures.

\subsection{Comparison to a Confidence-Only Baseline}
\label{sec:baseline}
 
Output disagreement costs more to compute than confidence: it needs the original FP32 model plus two forward passes per image (one FP32, one INT8), while confidence comes from the INT8 model's own output. We explore if this can be used a signal as well as decribed in Section~\ref{sec:method}. Table~\ref{tab:baseline} (Appendix~\ref{sec:baseline-appendix}) reports the result: Output disagreement separates accurate from inaccurate predictions more effectively than confidence, for all four architectures.

\section{Discussion}
\label{sec:discussion}
\label{sec:limitations}

We trained three convolutional architectures and one transformer-based architecture, and found that output cosine similarity correlates with accuracy degradation under INT8 quantization. In particular, below a threshold of 0.8, every architecture's performance fell below that of the baseline model. This suggests that as the outputs drift away from the baseline FP32 model, classification becomes less reliable. In a real-world setting without labels, this method could potentially serve as an uncertainty: providing clinicians with the percentage of samples that fall below this threshold, and letting them decide whether that level of uncertainty is acceptable for the given application before deployment. Future work will focus on exploring whether this signal holds for other datasets, as this paper has focused only on mammography data. Additionally, we will work toward improving the separation of cases where the FP32 model and its quantized counterpart disagree, thereby increasing confidence that a quantized model is making a mistake past a certain threshold and providing clinicians with a better label-free metric. Lastly, we will explore leveraging these unreliable cases to aid in training and improve accuracy for quantized models.
 

\section{Conclusion}
\label{sec:conclusion}

In this work, we investigated output cosine similarity as a label-free signal for detecting accuracy degradation under INT8 quantization, evaluating three convolutional architectures and one transformer-based architecture on mammography data. We show that this similarity correlates with accuracy degradation relative to the FP32 baseline.


\begin{ack}
\end{ack}

\bibliographystyle{plainnat}
\bibliography{references} 

\appendix

\section{Full Hyperparameters}
\label{sec:hyperparams}
\subsection{Training Details}
\label{sec:training-details}

The results reported in this paper use a single-stage (capacity-clamped) fine-tuning regimen on the CBIS-DDSM dataset \citep{Lee2017}, trained on ResNet-50 \citep{He2016}, VGG-16-BN \citep{Simonyan2014VeryDC}, Wide ResNet-50-2 \citep{DBLP:journals/corr/ZagoruykoK16}, and ViT-B/16 \citep{Dosovitskiy2021}. All models are initialized from ImageNet-1K pretrained weights. Table~\ref{tab:hyperparams} summarizes the core training hyperparameters; architecture-specific adaptations follow below.

\begin{table}[ht]
  \caption{Training hyperparameters for the single-stage fine-tuning regimen.}
  \label{tab:hyperparams}
  \centering
  \begin{tabular}{ll}
    \toprule
    \textbf{Hyperparameter} & \textbf{Value} \\
    \midrule
    Optimizer & Adam \\
    Learning rate & $5 \times 10^{-5}$ \\
    LR schedule & ReduceLROnPlateau (factor 0.5, patience 5) \\
    Weight decay & $1 \times 10^{-4}$ \\
    Batch size & 32 \\
    Max epochs & 30 \\
    Early stopping patience & 7 epochs \\
    Checkpoint selection & Best validation AUC \\
    Loss function & Cross-entropy \\
    Trainable parameters (ResNet-50) & 23.3M / 24.7M (94.2\%) \\
    \bottomrule
  \end{tabular}
\end{table}

\subsubsection{Training Regimen}
\label{sec:regimens}

The earliest backbone layers of each architecture are frozen for the entire training run, while subsequent layers and the classification head train jointly at a single fixed learning rate; the specific frozen layers per architecture are given in Section~\ref{sec:method}.

The classification head is a four-layer MLP ($2048 \to 512 \to 256 \to 128 \to 2$ for ResNet-50 and Wide ResNet-50-2; analogous dimensions adapted to each backbone's output dimensionality for VGG-16-BN and ViT-B/16) with ReLU activations and dropout ($p=0.2$) after each hidden layer.

\subsubsection{Data and Augmentation}

Training images are resized to $224 \times 224$ and enhanced with contrast-limited adaptive histogram equalization (CLAHE; clip limit 2.0, tile grid $8\times8$) prior to augmentation. Training-time augmentation consists of random horizontal and vertical flips ($p=0.5$ each), random rotation ($\pm15^\circ$), and Gaussian blur (kernel size $3\times3$, $\sigma \in [0.1, 2.0]$), with bilinear resizing and pixel normalization (mean/std $=0.5$). Validation and test images receive only resizing and normalization, no augmentation is applied at evaluation time.

\subsubsection{Statistical Significance}

To assess whether observed effects reflect consistent trends across initializations, we repeat all experiments across all four architectures over 3 random seeds: 1207, 804, and 510. We report mean $\pm$ standard deviation across these 3 seeds, computed per seed before pooling. Seeds control weight initialization of the classification head and data shuffling order; other sources of non-determinism are left at PyTorch defaults.

\subsubsection{Quantization}

Post-training static quantization is performed using PyTorch's FX graph mode quantization API (\texttt{prepare\_fx}/\texttt{convert\_fx}) with the \texttt{qnnpack} backend, targeting INT8 weights and activations. Calibration uses 500 images sampled from the training set (fixed seed 42) processed in evaluation mode to record activation statistics. \texttt{qnnpack} was selected as the quantization backend available in our ARM-based compute environment.

\subsubsection{Compute}

Experiments were run on an Apple M4 Pro system (48 GB unified memory) using PyTorch's MPS backend for training and CPU for INT8 quantization and evaluation. Training across 3 seeds for one architecture requires approximately 45 minutes. INT8 quantization and evaluation of a single checkpoint on the 704-image test set requires under 5 minutes on CPU. Feature-embeddings disagreement requires a single forward pass per precision per image, taking under 1 minute on CPU per seed.

\subsubsection{Feature-Embeddings Disagreement: Implementation Details}
\label{sec:outputvector-implementation}

For each test image and seed, we compute the softmax probability vector $p \in \mathbb{R}^2$ from both the FP32 model and the true converted INT8 model, recording cosine similarity alongside predicted class and correctness.

\textbf{Per-seed robustness analysis.} Threshold sensitivity is computed independently within each seed prior to aggregation to evaluate whether classification splits remain consistent across individual training runs.

\textbf{Confidence baseline comparison.} To compare confidence against feature-embeddings disagreement on a consistent basis, we apply an identical bottom-25th-percentile split ($n=288$ per architecture) to feature-embeddings cosine similarity.

\FloatBarrier

\section{Per-Similarity-Bin Accuracy: Full Detail}
\label{sec:extended-architectures}

Table~\ref{tab:allarch-bins} reports the full per-bin FP32-vs-INT8 accuracy underlying Figure~\ref{fig:binned-accuracy}, for all four evaluated architectures, computed per seed before pooling.

All four architectures show accuracy climbing with similarity, with the direction of the FP32-vs-INT8 gap consistent across bins. Minor INT8-exceeds-FP32 crossovers of $0.5$--$3.6$ points occur in the near-ceiling ($\geq0.90$ similarity) bins for VGG-16-BN and Wide ResNet-50, and $1.0$ points for ViT-B/16, consistent with sampling noise given FP32$\approx$INT8 by construction at that similarity level (ResNet-50 shows none).

\begin{table}[ht]
  \caption{Per-seed binned accuracy (FP32 vs.\ INT8), mean $\pm$ std across 3 seeds, output cosine similarity bins, all four evaluated architectures. Bins computed independently per seed before pooling.}
  \label{tab:allarch-bins}
  \centering
  \begin{tabular}{llcccc}
    \toprule
    \textbf{Architecture} & \textbf{Bin} & \textbf{FP32 Mean} & \textbf{FP32 Std} & \textbf{INT8 Mean} & \textbf{INT8 Std} \\
    \midrule
    ResNet-50 & $<0.80$ & $70.1$ & $7.2$ & $30.7$ & $8.6$ \\
    ResNet-50 & $0.80$--$0.90$ & $57.3$ & $17.4$ & $43.3$ & $5.0$ \\
    ResNet-50 & $0.90$--$0.95$ & $73.8$ & $2.1$ & $65.5$ & $6.0$ \\
    ResNet-50 & $0.95$--$1.00$ & $74.3$ & $3.0$ & $73.7$ & $1.5$ \\
    \midrule
    VGG-16-BN & $<0.80$ & $76.8$ & $20.8$ & $22.3$ & $20.4$ \\
    VGG-16-BN & $0.80$--$0.90$ & $62.8$ & $8.9$ & $59.5$ & $6.6$ \\
    VGG-16-BN & $0.90$--$0.95$ & $56.6$ & $4.4$ & $49.0$ & $6.6$ \\
    VGG-16-BN & $0.95$--$1.00$ & $74.1$ & $0.7$ & $74.6$ & $1.4$ \\
    \midrule
    Wide ResNet-50 & $<0.80$ & $52.4$ & $6.3$ & $44.6$ & $9.4$ \\
    Wide ResNet-50 & $0.80$--$0.90$ & $62.9$ & $6.5$ & $58.0$ & $2.6$ \\
    Wide ResNet-50 & $0.90$--$0.95$ & $58.3$ & $2.4$ & $61.9$ & $9.8$ \\
    Wide ResNet-50 & $0.95$--$1.00$ & $76.9$ & $2.7$ & $77.6$ & $3.6$ \\
    \midrule
    ViT-B/16 & $<0.80$ & $74.5$ & $0.9$ & $25.5$ & $0.9$ \\
    ViT-B/16 & $0.80$--$0.90$ & $69.3$ & $20.1$ & $50.5$ & $17.7$ \\
    ViT-B/16 & $0.90$--$0.95$ & $61.2$ & $8.3$ & $53.8$ & $10.1$ \\
    ViT-B/16 & $0.95$--$1.00$ & $73.5$ & $5.0$ & $74.5$ & $3.5$ \\
    \bottomrule
  \end{tabular}
\end{table}

\FloatBarrier

\section{Confidence Baseline Comparison: Full Table}
\label{sec:baseline-appendix}

Table~\ref{tab:baseline} reports the full comparison underlying Section~\ref{sec:baseline}: accuracy gap (bottom 25\% vs.\ top 75\% by signal value), both signals, fixed $n=288$ split per architecture.
 
\begin{table}[ht]
  \caption{Accuracy gap (bottom 25\% vs.\ top 75\%), confidence vs.\ output disagreement, fixed $n=288$ split. Bold marks the largest gap per architecture.}
  \label{tab:baseline}
  \centering
  \begin{tabular}{lcccc}
    \toprule
    \textbf{Architecture} & \textbf{Confidence} & \textbf{Output} \\
    \midrule
    ResNet-50 & $+17.4$ pts & $\mathbf{+30.3}$ \textbf{pts} \\
    VGG-16-BN & $+18.8$ pts & $\mathbf{+20.6}$ \textbf{pts} \\
    Wide ResNet-50 & $+15.4$ pts & $\mathbf{+27.9}$ \textbf{pts} \\
    ViT-B/16 & $+17.6$ pts & $\mathbf{+31.0}$ \textbf{pts} \\
    \bottomrule
  \end{tabular}
\end{table}

\section{Per-Seed Robustness: Full Detail}
\label{sec:perseed-appendix}

Table~\ref{tab:perseed} reports per-seed threshold sensitivity underlying Section~\ref{sec:outputvector}, for output disagreement, threshold 0.8. All four architectures produce reorganized cases in every seed; VGG-16-BN's smallest seed (1207, $n=4$, all misclassified) and ViT-B/16's smallest seed (1207, $n=4$) both reflect a lower baseline disagreement rate. Wide ResNet-50's per-seed accuracy-on-reorganized varies considerably ($36.1\%$--$54.7\%$), consistent with small per-seed sample sizes (40--86 reorganized cases) rather than a genuine inconsistency.

\begin{table}[H]
  \caption{Per-seed threshold sensitivity, output disagreement, threshold $=0.8$.}
  \label{tab:perseed}
  \centering
  \begin{tabular}{llccc}
    \toprule
    \textbf{Architecture} & \textbf{Seed} & \textbf{n (reorg.)} & \textbf{Acc.\ reorg.} & \textbf{Acc.\ stable} \\
    \midrule
    ResNet-50 & 510 & 52 & $23.1\%$ & $71.1\%$ \\
    ResNet-50 & 804 & 40 & $40.0\%$ & $71.2\%$ \\
    ResNet-50 & 1207 & 31 & $29.0\%$ & $70.3\%$ \\
    \midrule
    VGG-16-BN & 510 & 15 & $40.0\%$ & $72.4\%$ \\
    VGG-16-BN & 804 & 37 & $27.0\%$ & $71.8\%$ \\
    VGG-16-BN & 1207 & 4 & $0.0\%$ & $73.2\%$ \\
    \midrule
    Wide ResNet-50 & 510 & 64 & $54.7\%$ & $70.6\%$ \\
    Wide ResNet-50 & 804 & 61 & $36.1\%$ & $75.5\%$ \\
    Wide ResNet-50 & 1207 & 86 & $43.0\%$ & $75.5\%$ \\
    \midrule
    ViT-B/16 & 510 & 94 & $26.6\%$ & $73.8\%$ \\
    ViT-B/16 & 804 & 16 & $25.0\%$ & $65.5\%$ \\
    ViT-B/16 & 1207 & 4 & $25.0\%$ & $70.3\%$ \\
    \bottomrule
  \end{tabular}
\end{table}


\end{document}